\documentclass[letterpaper]{article}
\usepackage{aaai}
\usepackage{times}
\usepackage{helvet}
\usepackage{courier}
\usepackage{graphicx}
\usepackage{amsmath}
\usepackage{amssymb}
\usepackage{booktabs}
\usepackage{subfigure}
\usepackage{multirow}
\usepackage{algorithm}
\usepackage{algorithmic}

\begin{document}
%
\title{\Large Learning Category Correlations for Multi-label Image Recognition with Graph Networks}

\author{Qing~Li$^{1,2}$, \textbf{Xiaojiang~Peng}$^{2,*}$, \textbf{Yu~Qiao}$^{2}$, \textbf{Qiang~Peng}$^{1} $\\
$^{1}$ \text{Department of School of Information Science and Technology, Southwest Jiaotong University, Chengdu, China}\\
$^{2}$ \text{Shenzhen Institutes of Advanced Technology, Chinese Academy of Sciences, Shenzhen, China}\\
liqing1988@my.swjtu.edu.cn,~\{xj.peng,yu.qiao\}@siat.ac.cn
}
\maketitle
\begin{abstract}
\begin{quote}
Multi-label image recognition is a task that predicts a set of object labels in an image. As the objects co-occur in the physical world, it is desirable to model label dependencies. Previous existing methods resort to either recurrent networks or pre-defined label correlation graphs for this purpose.
In this paper, instead of using a pre-defined graph which is inflexible and may be sub-optimal for multi-label classification, we propose the A-GCN, which leverages the popular \textit{G}raph \textit{C}onvolutional \textit{N}etworks with an \textit{A}daptive label correlation graph to model label dependencies. Specifically, we introduce a plug-and-play Label Graph (LG) module to learn label correlations with word embeddings, and then utilize traditional GCN to map this graph into label-dependent object classifiers which are further applied to image features. The basic LG module incorporates two $1\times1$ convolutional layers and uses the dot product to generate label graphs. In addition, we propose a sparse correlation constraint to enhance the LG module, and also explore different LG architectures.  We validate our method on two diverse multi-label datasets: MS-COCO and Fashion550K. Experimental results show that our A-GCN significantly improves baseline methods and achieves performance superior or comparable to the state of the art.

\end{quote}
\end{abstract}

\section{Introduction}
As an important problem in computer vision community, multi-label image recognition has attracted considerable attention due to its wide applications such as music emotion categorization~\cite{trohidis2008multi}, fashion attribute recognition~\cite{inoue2017multi}, human attribute recognition~\cite{li2016human}, etc. Unlike conventional multi-class classification problems, which only predict one class label for each image, multi-label image recognition needs to assign multiple labels to a single image. Its challenges come from the rich and diverse semantic information in images.

Early existing methods~\cite{clare2001knowledge,tsoumakas2007multi,cheng2009combining,zhou2012multi,zhang2013review} address the multi-label classification problem by either transform it into i) multiple binary classification tasks or ii) multivariate regression problem or iii) adapting single-label classification algorithms.
With the great success of deep Convolutional Neural Networks (CNNs) on single-label multi-class image classification~\cite{krizhevsky2012imagenet}, recent multi-label image classification methods are mainly based CNNs with certain adaptions~\cite{wei2014cnn,wei2015hcp,wang2016cnn,wang2017multi,zhu2017learning,ge2018multi,chen2018order,yu2019delta,chen2019multi}.

A popular way of modern CNN-based multi-label classification is to model label dependencies as the objects usually co-occur in the physical world. For instance, 'baseball', 'bat' and 'person' always appear in the same image, but 'baseball' and 'ocean' rarely appear together.
Wang \textit{et al}.~\cite{wang2016cnn} propose a CNN-RNN framework, which learns a joint image-label embedding to characterize the semantic label dependency. It shows that the recurrent neural networks (RNNs) can capture the higher-order label dependencies in a sequential fashion. However, this method ignores the explicit associations between semantic labels and image regions. Consequently, some works combine the attention mechanism~\cite{xu2015show} with CNN-RNN framework to explore the associations between labels and image regions~\cite{wang2017multi,zhu2017learning,ge2018multi,chen2018order}. For example, Zhu \textit{et al}. \cite{zhu2017learning} propose a Spatial Regularization Network which generates class-related attention maps and captures both spatial and semantic label dependencies via learnable convolutions. These methods essentially learn local correlations by attention regions of an image which introduce limited complementary information. Chen \textit{et al}.~\cite{chen2019multi} propose a multilabel-GCN (ML-GCN) framework, which leverages Graph Convolutional Networks to capture global correlations between labels with extra knowledge from label statistical information. One drawback of ML-GCN is that the label correlation graph is manually designed and needs carefully adaptions. This hand-crafted correlation graph makes the ML-GCN inflexible and may be sub-optimal for multi-label classification. 

In this paper, we propose a unified multi-label GCN framework, termed as A-GCN to address the inflexible correlation graph problem in ML-GCN.  The key of A-GCN is that it learns an \textit{A}daptive label correlation graph to model label dependencies in an end-to-end manner. Specifically, we introduce a plug-and-play \textit{adaptive Label Graph (LG) module} to learn label correlations with word embeddings, and then utilize traditional GCN to map this graph into label-dependent object classifiers, and further applied these classifiers to image features. By default, we implement LG module by two $1\times1$ convolutional layers and uses dot product to generate label graphs. As label co-occurance is sparse in current popular multi-label datasets, we also introduce a \textit{sparse correlation constraint} to enhance the LG module by using a L1-norm loss between the learned correlation graph and an identity matrix. Furthermore, we explore three alternative architectures to evaluate the LG module.
We validate our method on two diverse multi-label datasets: MS-COCO and Fashion550K. Experimental results show that our A-GCN significantly improves baseline methods and achieves performance superior or comparable to the state of the art.

\begin{figure*}[tp]
\centering{
  \includegraphics[width=0.8\linewidth]{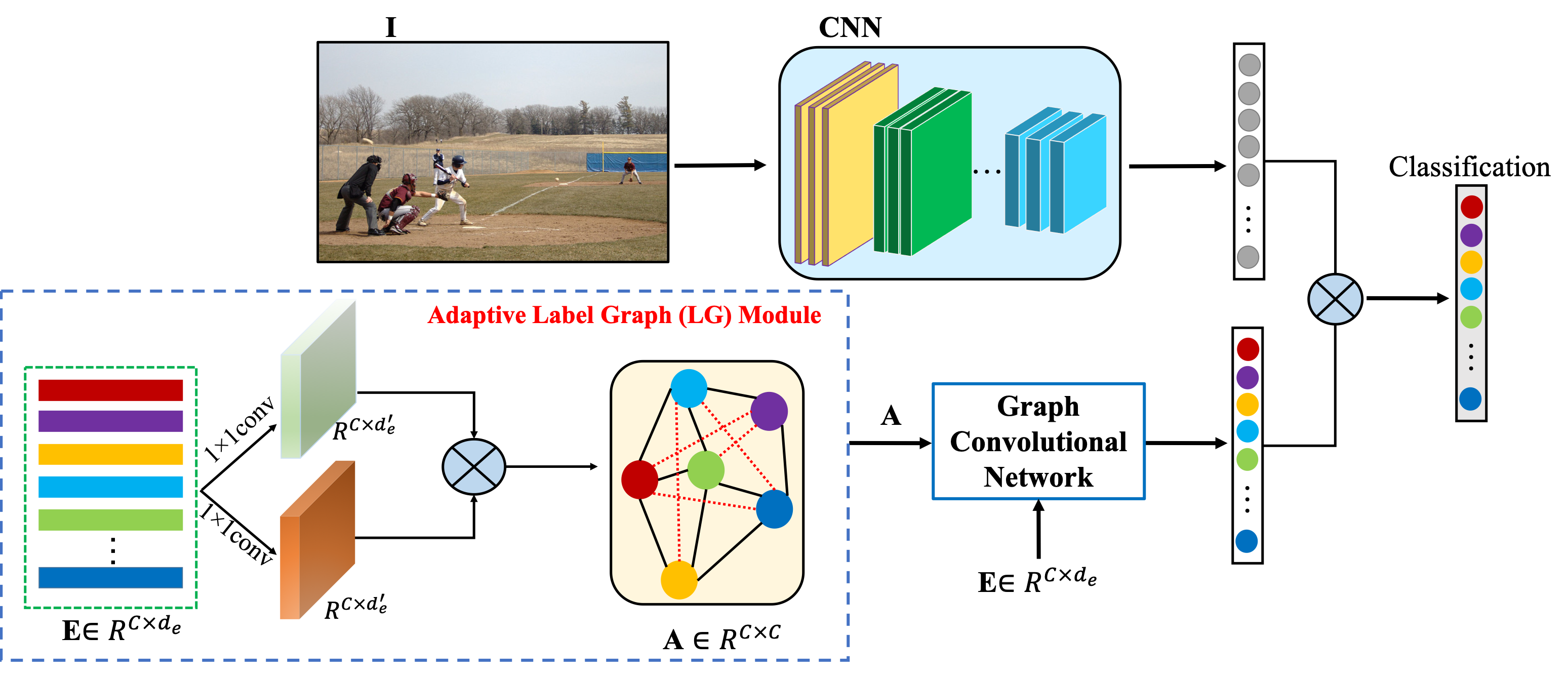}\\
  \caption{The pipeline of our A-GCN for multi-label image recognition. It consists of two branches, namely an image-level branch to extract image features and a label GCN branch to learn label-dependent classifiers. An adaptive \textit{label graph} (LG) module is introduced to construct the label correlation matrix from label embeddings for the label GCN branch.}\label{fig:figure1}
  }
\end{figure*}

\begin{figure}
\center
\includegraphics[width=0.9\linewidth]{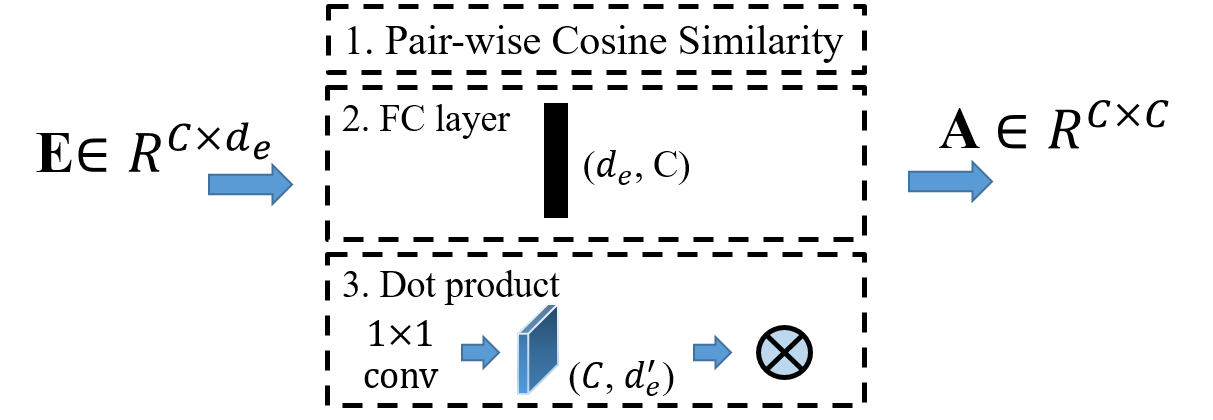}
\caption{Three kinds of alternative label graph architectures.}
\label{fig:lg}
\end{figure}

\section{Relate Work}
Our work is mainly related to multi-label image recognition and graph neural network. In this section, we first present related works on multi-label image recognition methods, and then graph neural network methods.

\subsection{Multi-label Image Recognition}

Remarkable developments in image recognition have been observed over the past few years due to the availability of large-scale hand-labeled datasets like ImageNet \cite{deng2009imagenet} and MS-COCO \cite{lin2014microsoft}. Recent progress on single-label image classification is made based on the deep convolutional neural networks (CNNs) \cite{krizhevsky2012imagenet,simonyan2014very,he2016deep} that learn powerful visual representation via stacking multiple nonlinear transformations. A simple way is to adapt these single-label classification networks to the multi-label image recognition with the deep CNNs, which has been witnessed good results \cite{sharif2014cnn,wang2016cnn,wang2017multi,chen2018recurrent}.

Early works on multi-label image recognition utilize hand-crafted image features and linear models to solve this problem \cite{tsoumakas2009mining,tai2012multilabel,cabral2014matrix,chen2012hierarchical}. Intuitively, as a well-known example is to decompose the multi-label image recognition problems into multiple binary classification tasks \cite{tsoumakas2007multi}. As in paper \cite{tsoumakas2009mining}, to train a set of independent linear classifiers for each label. Zhang et al. \cite{zhang2007ml} propose a multi-label lazy learning approach named ML-KNN, using k-nearest neighbor to predict labels for unseen data from training data. Tai et al. \cite{tai2012multilabel} design a novel Principle Label Space Transformation (PLST) algorithm,  which seeks important correlations between labels before learning. Chen et al. \cite{chen2012hierarchical} introduce a hierarchical matching framework with so-called side information for image classification based on the bag-of-words model. Although these methods may perform well on the simple benchmarks, they can't generalize as well as deep learning-based methods on input images with complex scenes and multiple objects.

Several studies based on CNNs still attract the attention of researchers in Multi-label image recognition tasks \cite{chen2012hierarchical,wang2016cnn,wang2017multi}. The earliest applications of deep learning to multi-label classification is done by Gong et al. \cite{gong2013deep}, who propose to combine convolutional architectures with an approximate top-k ranking objective function for annotating multi-label images. Instead of extracting off-the-shelf deep features, Chatfield et al. \cite{chatfield2014return} fine-tune the network using the target multi-label dataset, which is used to learn more domain-specific features to boost the classification performance. Wu et al. \cite{wu2015weakly} propose an approach named weakly semi-supervised deep learning for multi-label image annotation, which uses a triplet loss function to draw images with similar label sets. To better consider the correlations between labels instead of treat each label independently, various approaches have been considered in recent works. One of the popular trends utilizes the graph models to build the label co-occurrence dependency \cite{tehrani2017modeling}, such as Conditional Random Field \cite{ghamrawi2005collective}, Dependency Network \cite{guo2011multi}, and Co-occurrence Matrix \cite{xue2011correlative}. In order to explore the label co-occurrence dependency combined with CNNs model, another group of researchers applies the low-dimensional recurrent neurons in RNN model to efficiently abstract high-order label correlation.
For example, Wang et al. \cite{wang2016cnn} utilize the RNNs combined with CNN to learn a joint image-label embedding to characterize the semantic label dependency as well as the image-label relevance. Wang et al. \cite{wang2017multi} introduce a spatial transformer layer and long short-term memory (LSTM) units to capture the label correlation. Lee et al. \cite{lee2018multi} propose a framework that incorporates knowledge graphs for describing the relationships between multiple labels and learned representations of this graph to enhance image feature representation to promote multi-label recognition.

\subsection{Graph Convolutional Neural Networks}
Generalization of GCNNs has drawn great attention in recent years. There are two typical types of GCNNs: spatial manner and spectral manner. The first type adopts feed-forward neural networks to every node \cite{scarselli2008graph}. For example, Marino et al. \cite{marino2016more} successfully apply GCNNs for multi-label image classification to exploit explicit semantic relations in the form of structured knowledge graphs. Wang et al. \cite{wang2018videos} propose to represent videos as space-time region graphs which capture similarity relationships and spatial-temporal relationships. Wang et al. \cite{wang2019linkage} propose a spatial-based GCN to solve the link prediction problem. The second type provides well-defined localization operators on graphs via convolutions in the Fourier domain \cite{kipf2016semi}. In recent years, an important branch of the spectral GCNNs has been proposed to tackle graph-structured data. The outputs of spectral GCNNs are updated features for each object node, leading to an advanced performance on any tasks related to graph based information processing. More specifically, Kipf et al. \cite{kipf2016semi} apply the GCNNs to semi-supervised classification. Hamilton et al. \cite{hamilton2017inductive} leverage GCNs to learn feature representations. \cite{chen2019multi} propose a novel GCN based model (aka ML-GCN) to learn the label correlations for multi-label image recognition tasks. It utilizes the GCN to learn an object classifier via mining their co-occurrence patterns within the dataset. Motivated by ML-GCN \cite{chen2019multi}, our work leverages the graph structure to capture and explore an adaptive label correlation graph. With the proposed A-GCN, we can overcome the limitation caused by the manually designed graph and automatically learn the label correlation by an LG module. We also demonstrate that our A-GCN is also an effective model for label dependency and can be trained in an end-to-end manner.

\renewcommand{\algorithmicrequire}{ \textbf{Input:}} 
\renewcommand{\algorithmicensure}{ \textbf{Output:}} 
\begin{algorithm}[htb]
\caption{Training of A-GCN}
\label{alg:auto-gcn}
\begin{algorithmic}
\REQUIRE ~~\\
image data and ground-truth label data ($\mathbf{I}$, $\mathbf{Y}$);\\
labels' word embedding $E$;\\
\ENSURE ~~\\
image-level feature $\mathbf{X}$;\\
adaptive Correlation Matrix $\mathbf{A}$;\\
label-dependent classifiers $\bar{\mathbf{W}}$;\\
the final predicted score vector $\mathbf{P}$;\\
\STATE Repeat:
\STATE \quad\textbf{Branch 1: Feedforward of image CNN}
\STATE  \qquad Extract image feature X : \\
\STATE  \qquad $X = f_{CNN}(I; \theta_{CNN})$; \\
\STATE \quad\textbf{Branch 2: Feedforward of label-dependent classifiers}
\STATE \qquad Learn/initialize the correlation matrix A with \\
\STATE \qquad labels' word embedding:\\
\STATE $\qquad A \gets Eq.(\ref{eq:A})$;
\STATE \qquad Compute the $L_{A}$: \\
\STATE $\qquad L_{A} \gets Eq.(\ref{eq:L1})$;
\STATE \qquad Learn the label-dependent classifier $\bar{\mathbf{W}}$ by  GCN:\\
\STATE \qquad $\bar{W} = f_{GCN}(E,\widehat{A}; \theta_{GCN}) \gets Eq.(\ref{eq:gcn}$;
\STATE \qquad Get predictions by applying classifier $\bar{\mathbf{W}}$ to
\STATE \qquad image feature X:\\
\STATE $\qquad P \gets Eq.(\ref{eq:p})$;
\STATE \qquad Compute the $L_{classifier}$:\\
\STATE $\qquad L_{classifier}(P;Y) \gets Eq.(\ref{eq:msml})$;
\STATE \qquad Compute the $L_{total}$: \\
\STATE $\qquad L_{total} = L_{classifier} + \alpha* L_{A}$;
\STATE Backpropagation until $L_{total}$ converges;
\end{algorithmic}
\end{algorithm}

\section{Approach}
To efficiently exploit the label dependencies and make GCN flexible, we propose the A-GCN to learn label correlations for GCN based multi-label image classification. In this section, we first present some notations to define the problem, and then introduce the basic GCN based multi-label classification, finally we present our A-GCN and several alternative label graph architectures.

\subsection{Preliminaries}
\textbf{Notations}.
Let $D=\{( \mathbf{I}_{i},\mathbf{y}_{i}) \mid i = 1 \ldots N \}$ be the training data, where $ \mathbf{I}_{i}$ is the $i-$th image and $\mathbf{y}_{i} \in \left\{ 0,1 \right\}^{C}$ is the corresponding multi-hot label vector. Zeros or ones in the label vector $\mathbf{y}$ denote the absence or presence of the corresponding category in the image. Let $\mathbf{x}_i \in R^{D}=f(\mathbf{I}_{i}; \theta)$ denote the CNN feature of $ \mathbf{I}_{i}$, and $f(\cdot; \theta)$ as a CNN model with parameters $\theta$.
Assume we have object classifiers $\bar{\mathbf{W}} \in R^{C\times D}=\{\mathbf{\bar{w}}_i\}^C_{i=1}$, then the predicted logit scores of feature $\mathbf{x}_i$ can be defined as,

\begin{equation}\label{eq:p}
\mathbf{p}_i=\bar{\mathbf{W}}\mathbf{x}_i.
\end{equation}

The CNN model and classifiers can be optimized by the following multi-label classification loss,

\begin{equation}\label{eq:msml}
    L_{classifier} = -\frac{1}{C} \sum_{j=1}^{C}\mathbf{y}_{i}^j\log{(\sigma(\mathbf{p}_{i}^j))} +(1-\mathbf{y}_{i}^j)*\log{(1-\sigma(\mathbf{p}_{i}^j))}
\end{equation}
where $\sigma(\cdot)$ the sigmoid function.

\textbf{Multi-label classification with GCN}.
We revisit the ML-GCN~\cite{chen2019multi} pipeline for multi-label classification in the following. It performs GCN on the word embeddings $\mathbf{E}\in R^{C\times d_e}$ of labels, and learns inter-dependent object classifiers to improve performance.
The purpose of GCN is to learn a function on a graph $\mathcal{G} = (\mathcal{V},\mathcal{E}) $, which takes previous feature descriptions $\mathbf{H}^l\in R^{C\times d}$ and the correlation matrix $\mathbf{A}\in R^{C\times C}$, and outputs learned node features as  $\mathbf{H}^{l+1}\in R^{C\times d^{'}}$. One GCN layer can be formulated as,

\begin{equation}\label{eq:gcn}
\centering
\mathbf{H}^{(l+1)} = \delta (\widehat{\mathbf{A}}\mathbf{H}^{(l)}\mathbf{W}^{(l)}),
\end{equation}

where
\begin{equation}\label{eq:Ahat}
\widehat{\mathbf{A}} = \widetilde{D}^{-\frac{1}{2}}(\mathbf{A}+\mathbf{I}_{C})\widetilde{D}^{-\frac{1}{2}},
\end{equation}
and $\mathbf{W}^{(l)}\in R^{d\times d^{'}}$ is a transformation matrix to be learned, $\widehat{\mathbf{A}}$ is the normalized version of $\mathbf{A}$ with $\widetilde{D}_{ii} = \sum_{j}\widetilde{\mathbf{A}}_{ij}$ and $\widetilde{\mathbf{A}}=\mathbf{A}+\mathbf{I}_{C}$,  $\mathbf{I}_{C}$ is an identity matrix, and $\delta(\cdot)$ is an activation function which is set as LeakyReLU following \cite{chen2019multi}.
The input of the first layer is $\mathbf{E}$ and the output of the last layer is $\bar{\mathbf{W}} \in R^{C\times D}$, i.e. the inter-dependent classifiers.

The crucial problem of ML-GCN is how to build correlation matrix $\mathbf{A}$. \cite{chen2019multi} constructs it via mining label co-occurance within the target datasets. To overcome the over-smoothing problem of $\mathbf{A}$, it either binarizes or re-weights the original co-occurance matrix with thresholding.

\subsection{A-GCN}
Following the pipeline of ML-GCN, we propose the A-GCN to address the generation of label correlation matrix $\mathbf{A}$.
Figure \ref{fig:figure1} depicts the framework of A-GCN. It mainly consists of two branches. The upper branch is a traditional CNN for image feature learning, and the bottom branch is a GCN model to generate inter-dependent classifiers.

The key difference between our A-GCN and ML-GCN is the construction method of $\mathbf{A}$. We argue that building correlation matrix $\mathbf{A}$ by counting the occurrence of label pairs and thresholding is inflexible and may be sub-optimal for multi-label classification. To address this problem, we propose an adaptive label graph (i.e correlation matrix) module to learn label correlations in an end-to-end manner.

\textbf{Adaptive label graph (LG) module}.
As shown in Figure \ref{fig:figure1}, the adaptive LG module is comprised of two $1\times 1$ convolutional layers and a dot product operation. The LG module takes as input the word embeddings of labels and output a learned label correlation matrix $\mathbf{A}$. Formally, the learned $\mathbf{A}$ can be written as,

\begin{equation}\label{eq:A}
\mathbf{A} = \frac{1}{C}(\mathbf{W}_{\phi}*\mathbf{E})^{T}(\mathbf{W}_{\theta}*\mathbf{E})
\end{equation}
where $ W_{\phi}$ and $W_{\theta}$ are the convolutional kernels to be learned, and $*$ denotes the convolutional operation.

Following the normalization trick in \cite{kipf2016semi}, we normalize $\mathbf{A}$ to $\widehat{\mathbf{A}}$ by Equation (\ref{eq:Ahat}).

\textbf{Sparse correlation constraint}. For each node of a certain graph, GCN gradually aggregates information from its own feature and the adjacent nodes' features. We can imagine that the features can be indistinguishable by over-smoothing if the learned $\mathbf{A}$ becomes uniform. A uniform $\mathbf{A}$ denotes dense correlations among different labels. To avoid this issue, we enforce a sparse correlation constraint on  $\widehat{\mathbf{A}}$ by a L1-norm loss as follows,
\begin{equation}\label{eq:L1}
L_{A} =  |\widehat{\mathbf{A}} - \mathbf{I}_{C}  |.
\end{equation}
This constraint encourages high self-correlation weights to avoid over-smoothed features in GCN. Our total loss is $L_{total} = L_{classifier} + \alpha*L_{A}$, where $\alpha$ is a trade-off weight and is default as 1.0 in our experiments.

\textbf{Alternative LG architectures}. As illustrated in Figure \ref{fig:lg}, we propose three alternative LG architectures, namely i) pair-wise cosine similarity (abbreviated as \textbf{Cos-A}), ii) linear transformation of $\mathbf{E}$ by a full-connected layer (\textbf{FC-A}), and iii) linear transformation of $\mathbf{E}$ with a dot product (\textbf{Dot-A}).

\textbf{Cos-A} simply computes the cosine similarities between label embeddings which generates a symmetrical correlation matrix. Each element in $\mathbf{A}$ is defined by,
\begin{equation}\label{eq:A1}
\mathbf{A}(i,j) = \cos(\mathbf{E}_{i},\mathbf{E}_{j}).
\end{equation}

\textbf{FC-A} directly utilizes a linear layer $\mathbf{W}_l\in R^{d_e\times C}$ to generate the correlation matrix as,
\begin{equation}\label{eq:A2}
\mathbf{A}= \mathbf{W}_l^\top~\mathbf{E}.
\end{equation}

\textbf{Dot-A} first uses a $1\times 1$ convolutional layer for $\mathbf{E}$ and a dot product operation, and then compute the self-correlation matrix as $\mathbf{A}$,
\begin{equation}\label{eq:A3}
\mathbf{A}= \frac{1}{C}(\mathbf{W}_{\phi}*\mathbf{E})^{T}(\mathbf{W}_{\phi}*\mathbf{E})
\end{equation}

\begin{table*}[tp]
\centering
\caption{Performance comparison of our framework and state-of-the-art methods on MS-COCO. $^\ast$It denotes our re-implementation results.}\label{tab:table2}
\scalebox{0.9}{
\begin{tabular}{l|c|c|c|c|c|c|c|c|c|c|c|c|c}
\toprule
 \multicolumn{1}{c|}{\multirow{2}*{Model}} &  \multicolumn{7}{c|}{All} & \multicolumn{6}{c}{Top-3} \\
 \cline{2-14}
  \multicolumn{1}{c|}{} & mAP & CP & CR & CF1 & OP & OR & OF1 & CP & CR & CF1 & OP & OR & OF1\\
 \hline
 CNN-RNN & 61.2 & - & - & - & - & - & - & 66.0 & 55.6 & 60.4 & 69.2 & 66.4 & 67.8 \\
 RNN-Attention & - & - & - & - & - & - & - & 79.1 & 58.7 & 67.4 & 84.0 & 63.0 & 72.08 \\
 Order-Free RNN & - & - & - & - & - & - & - & 71.6 & 54.8 & 62.1 & 74.2 & 62.2 & 67.7 \\
 ML-ZSL & - & - & - & - & - & - & - & 74.1 & 64.5 & 69.0 & - & - & - \\
 SRN & 77.1 & 81.6 & 65.4 & 71.2 & 82.7 & 69.9 & 75.8 & 85.2 & 58.8 & 67.4 & 87.4 & 62.5 & 72.9 \\
 Multi-Evidence & - & 80.4 & 70.2 & 74.9 & 85.2 & 72.5 & 78.4 & 84.5 & 62.2 & 70.6 & 89.1 & 64.3 & 74.7 \\
  ML-GCN (Binary) & 80.3 & 81.1 & 70.1 & 75.2 & 83.8 & 74.2 & 78.7 &  84.9 &  61.3 & 71.2 & 88.8 & 65.2 & 75.2 \\
 ML-GCN (Re-weighted) & 83.0 & \textbf{85.1} & 72.0 & \textbf{78.0} & 85.8 & 75.4 & \textbf{80.3} & \textbf{89.2} & 64.1 & \textbf{74.6} & \textbf{90.5} & 66.5 & \textbf{76.7} \\
 ML-GCN (Re-weighted)$^{\ast}$ & 82.5 & 83.7 & 72.0 & 77.4 & 84.7 & 75.5 & 79.8 & 88.4 & 63.8 & 74.1 & 89.9 & 66.2 & 76.3 \\
 Our baseline (ResNet101) & 80.3 & 77.8 & 72.8 & 75.2 & 81.5 & 75.1 & 78.2 & 82.5 &  64.6  &  72.4  & 87.3 &  65.7  & 75.0 \\
\hline\hline
  A-GCN  & \textbf{83.1} & 84.7 & 72.3 & \textbf{78.0} & 85.6 & 75.5 & \textbf{80.3} & 89.0 & 64.2 & \textbf{74.6} & \textbf{90.5} & 66.3 & 76.6 \\
 \hline\hline
 A-GCN (w/o $L_A$) & 82.78 & 83.04 & \textbf{72.87} & 77.63 & 84.45 & \textbf{75.75} & 79.87 & 87.48 & \textbf{64.73} & 74.4 & 89.55 & \textbf{66.54} & 76.35 \\
 Cos-A (w $L_A$) & 82.77 & 84.89 & 71.67 & 77.72 & 85.77 & 74.83 & 79.93 & 88.92 & 64.03 & 74.45 & 90.24 & 66.2 & 76.37 \\
 FC-A  (w $L_A$) & 82.85 & 83.65 & 72.45 & 77.65 & 84.99 & 75.56 & 80.0 & 88.29 & 64.23 & 74.37 & 89.95 & 66.3 & 76.34 \\
 Dot-A  (w $L_A$) & 82.22 & 84.64 & 70.93 & 77.18 & \textbf{85.86} & 74.65 & 79.87 & 88.74 & 63.19 & 73.82 & 90.37 & 65.93 & 76.24 \\
\bottomrule
\end{tabular}}
\end{table*}

\textbf{Training}. We illustrate the training process of A-GCN in Algorithm 1. We train A-GCN in an end-to-end manner with two branches. \textbf{Branch 1} extracts image features and updates image-level CNN parameters. \textbf{Branch 2} learns adaptive label correlation graph and the GCN model to generate label-dependent classifiers. The total loss is the combination of sparse correlation constraint $L_{A}$ and multi-label classification loss $L_{classifier}$.


\section{Experiment}
In this section, we evaluate the proposed A-GCN and compare it to the state-of-the-art methods on two public multi-label benchmark datasets: MS-COCO \cite{lin2014microsoft} and Fashion550K \cite{inoue2017multi}. We first present the implementation details and metrics, and then extensively explore our A-GCN on MS-COCO, and finally apply A-GCN on Fashion550K.

\subsection{Implementations and evaluation metrics}
We implement our method with Pytorch. For data augmentation, we resize images to scale 512$\times$512 on MS-COCO (256$\times$256 on Fashion550K), and randomly crop regions of 448$\times$448 (224$\times$224 on Fashion550K) with random flipping. For test, we resize images to scale 448$\times$448 (224$\times$224).
For fair comparison, we use ResNet-101 on MS-COCO \cite{chen2019multi}, and ResNet-50 on Fashion550K \cite{inoue2017multi}, which are pre-trained on ImageNet. We use the SGD method for optimization with a momentum of 0.9 and a weight decay of $10^{-4}$. We set the minibatch size as 32, the initial learning rate ($lr$) as $10^{-2}$. We divide $lr$ by 10 after every 30 epochs, and stop training after 65 epochs. For word embedding method and other hyper-parameters of GCN are kept consistent with \cite{chen2019multi}.

\textbf{Evaluation metrics}.
For MS-COCO dataset, we use the same evaluation metrics as \cite{chen2019multi}, i.e. the mean of class-average precision (\textbf{mAP}), overall precision (OP), recall (OR), F1 (OF1), and average per-class precision (CP), recall (CR), F1 (CF1). For each image, the labels are predicted as positive if the confidences of them are greater than 0.5. Among all these metrics, mAP is known as the most important one. For fair comparisons, we also report the results of top-3 labels.%
On Fashion550K, we also use \textbf{mAP} and the class agnostic average precision ($AP_{all}$) to evaluate the performance for consistency with \cite{inoue2017multi}.

\begin{figure}
\centering
\includegraphics[width=0.8\linewidth]{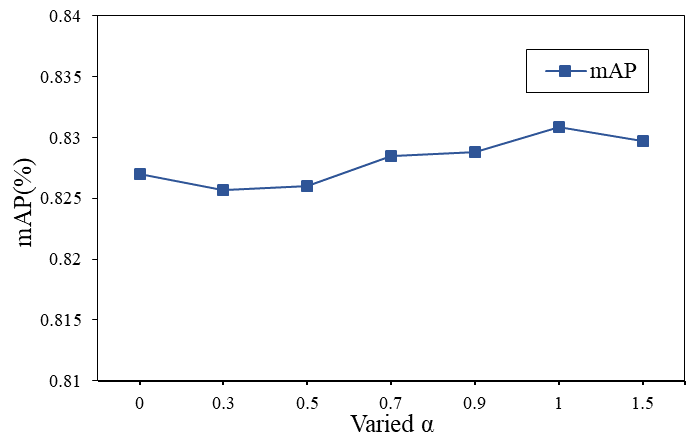}
\caption{Accuracy comparisons with different values of $\alpha$.}
\label{fig:alpha}
\end{figure}

\begin{figure*}[htp]
\centering
\subfigure[Baseline vs A-GCN on MS-COCO dataset]
{
\includegraphics[width=0.45\linewidth]{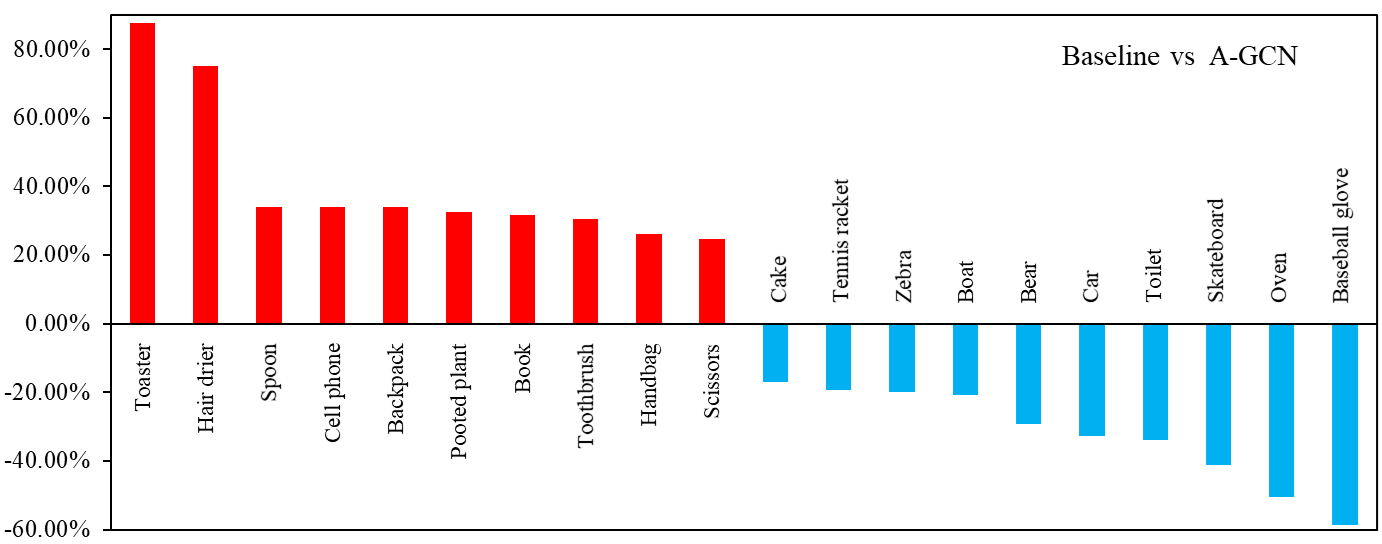}
}
\subfigure[ML-GCN vs A-GCN MS-COCO dataset]
{
\includegraphics[width=0.45\linewidth]{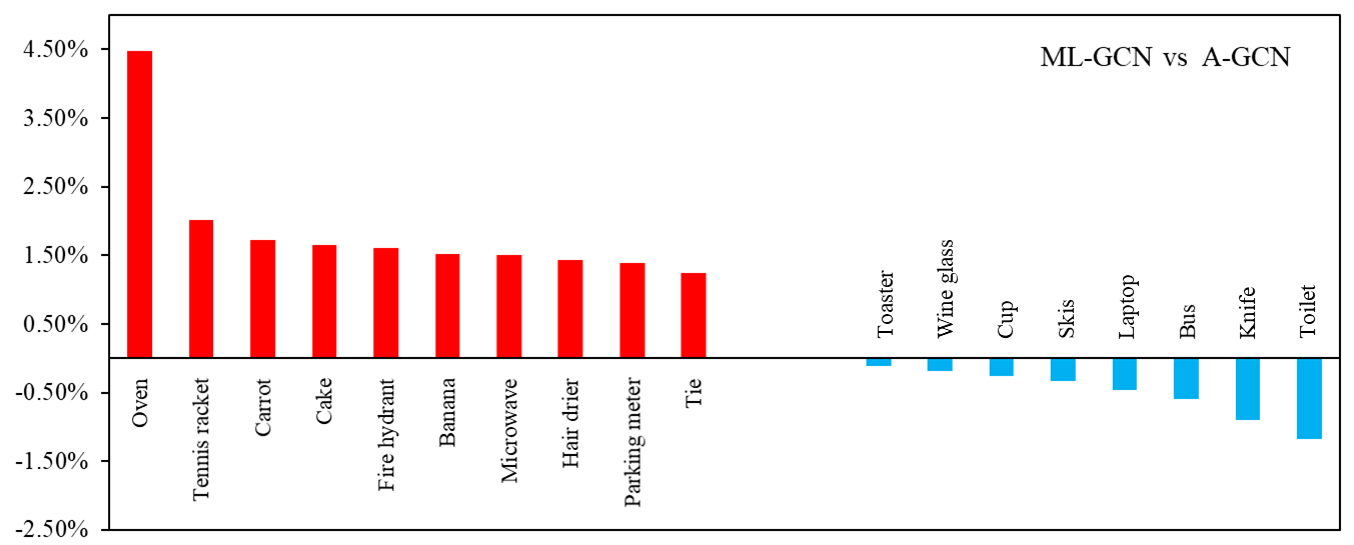}
}
\quad
\subfigure[Baseline vs A-GCN on Fashion550K dataset]
{
\includegraphics[width=0.45\linewidth]{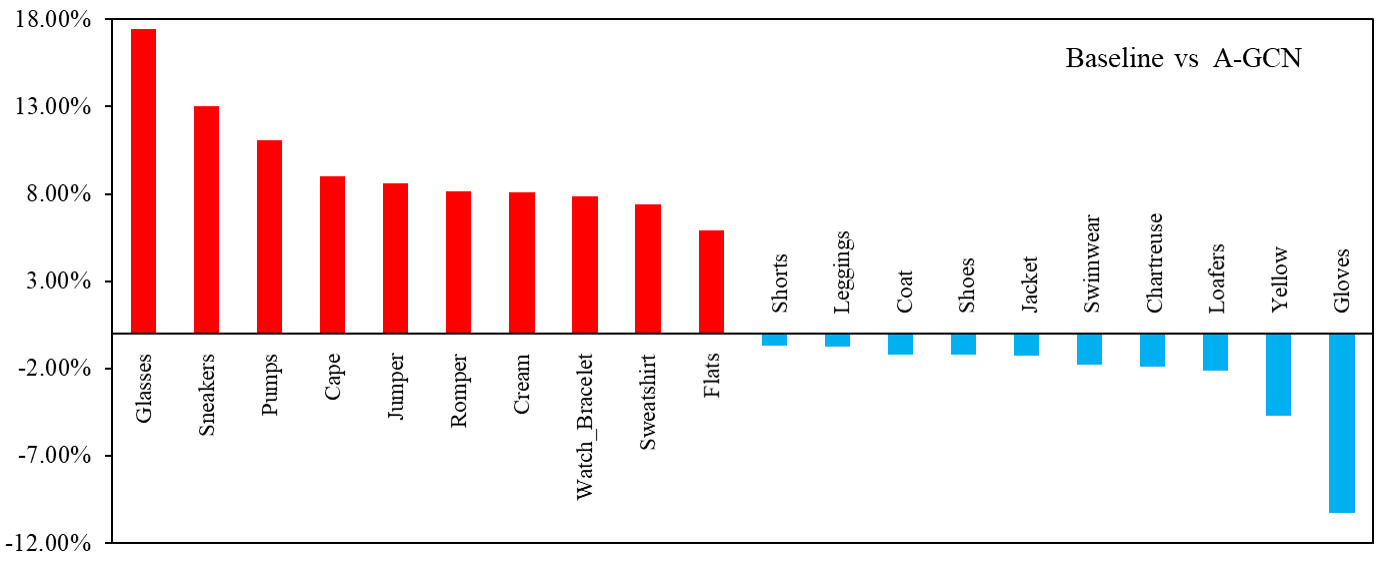}
}
\subfigure[ML-GCN vs A-GCN on Fashion550K dataset]
{
\includegraphics[width=0.45\linewidth]{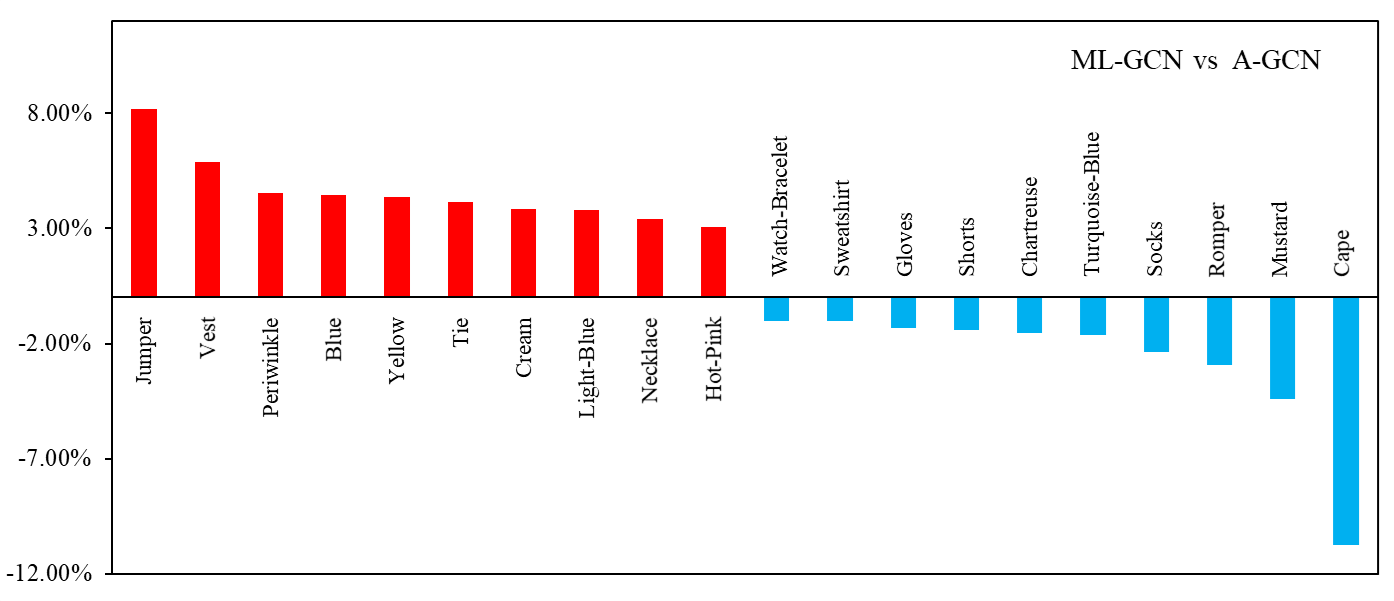}
}
\caption{Per-class improvement or degradation of AP between A-GCN and (or ML-GCN) on MS-COCO (or Fashion550k). The top-10 improved classes from our A-GCN are indicated as red, and the top-10 degraded classes blue.}
\label{fig:tabel_A}
\end{figure*}

\subsection{Exploration on MS-COCO}
MS-COCO is the most popular multi-label image dataset which consists of 80 categories with 82,081 training images and 40,137 test images.
We compare our A-GCN to several state-of-the-art methods including CNN-RNN~\cite{wang2016cnn}, RNN-Attention~\cite{wang2017multi}, Order-Free RNN~\cite{chen2018order}, ML-ZSL~\cite{lee2018multi}, SRN~\cite{zhu2017learning}, Multi-Evidence~\cite{ge2018multi}, ML-GCN~\cite{chen2019multi}. The results are presented in Table \ref{tab:table2}. Our A-GCN significantly improves the baseline (ResNet101) in most of the metrics. Specifically, the A-GCN improves the mAP of baseline from 80.3\% to 83.1\%. In addition, our A-GCN slightly outperforms the most related method ML-GCN in mAP. Compared to ML-GCN, our A-GCN, with a small extra LG module, is more flexible which does not need to elaborately design correlation matrix.

\textbf{Evaluation of $L_A$ and LG architectures}.
We evaluate the effect of sparse correlation constraint $L_A$ and different label graph architectures in the last four rows of Table \ref{tab:table2}. Several observations can be concluded as following. First, without $L_A$ we obtain slightly worse results than the default A-GCN, which indicates the effectiveness of sparse constraint. Second, all alternative LG architectures improve the baseline obviously which suggests that all of them learn label correlation information effectively. Third, the \textbf{FC-A}, which only differs from the default A-GCN by replacing $1\times 1$ convolutional with one FC layer, shows the best results in all the alternative ones. Compared to the default A-GCN, the \textbf{Dot-A} has an obviously degradation.

\textbf{Evaluation of $\alpha$}.
The trade-off weight $\alpha$ indicates the contribution of $L_A$ in the whole loss value. Intuitively, this regularization should not have large weight. Figure \ref{fig:alpha} shows the evaluation of $\alpha$ on MS-COCO. Increasing $\alpha$ from 0 to 1 slightly boosts performance, while larger $\alpha$ leads to degradation and even divergence ($\alpha=2.0$ in our test).

\textbf{Visualization}.
To further investigate the effect of our A-GCN, we show the per-class improvement (degradation) from A-GCN on MS-COCO and Fashion550K in Figure \ref{fig:tabel_A}. It shows that those objects (mainly daily needs) whose presences usually depend on their co-occurrence containers are likely to have large gains, e.g. spoon, backpack,
book, toothbrush in image (a), (or glasses, sneakers, sweatshirts in image (c)), etc.
It suggests that our A-GCN leverages the graph module to automatically learn the objects co-occurrence relation, which can effectively improve the multi-label recognition performance.

\subsection{Performance on Fashion550K}
Fashion550K~\cite{inoue2017multi} is a multi-label fashion dataset which contains 66 unique weakly-annotated tags with 407,772 images in total. Among all the images, 3,000 images are manually verified for training (i.e. \textbf{clean} set), 300 images for validation, and 2,000 images for test. The rest images are used as noisy-labeled data, i.e. \textbf{noisy} set. We report performance on the test set following common setting.

We compare our default A-GCN to several well-known state-of-the-art methods on Fashion550K, including StyleNet~\cite{simo2016fashion}, Baseline and Inoue et al. proposed method \cite{inoue2017multi}, Viet et al. proposed method \cite{veit2017learning}, and our re-implementation ML-GCN (Re-weighted). For fair comparison, we also use two training configurations, namely i) training on noisy set and ii) further fine-tuning on clean set (i.e. noisy+clean). The comparison is presented in Table \ref{tab:table1}. 
Our A-GCN improves our baseline by 2.76\% and 3.4\% in mAP with both training settings, respectively.
It also demonstrates the label correlation information is helpful for multi-label fashion image classification.

\begin{table}[tp]
\centering
\caption{Comparison of $AP_{all}$ and $mAP$ on Fashion550K.}\label{tab:table1}
\begin{tabular}{lllc}
\toprule
  Model & Data &$ AP_{all} $& mAP \\
 \hline
  Baseline & noisy & 69.18 $\%$ & 58.68 $\%$ \\
  StyleNet & noisy & 69.53 $\%$ & 53.24 $\%$ \\
  ML-GCN & noisy & 68.46 $\%$ & 60.85 $\%$ \\
  Our baseline & noisy & 68.26 $\%$ & 58.59 $\%$ \\
  \hline
   A-GCN & noisy & \textbf{70.28} $\%$ & \textbf{61.35} $\%$ \\
  \hline
  \hline
  Baseline & noisy+clean & 79.39 $\%$ & 64.04 $\%$ \\
  Viet et al. & noisy+clean & 78.92 $\%$ & 63.08 $\%$ \\
  Inoue et al. & noisy+clean & 79.87 $\%$ & 64.62 $\%$ \\
  ML-GCN & noisy+clean & 80.52 $\%$ & 65.74 $\%$ \\
  Our baseline & noisy+clean & 77.84 $\%$ & 62.92 $\%$ \\
  \hline
   A-GCN & noisy+clean & \textbf{80.95} $\%$ & \textbf{66.32} $\%$ \\
\bottomrule
\end{tabular}
\end{table}

\section{Conclusion}
In this paper, we proposed a simple and flexible A-GCN framework for multi-label image recognition. The A-GCN leverages a plug-and-play label graph module to automatically construct the label correlation matrix for GCN on the label embeddings. We designed a sparse correlation constraint on the learned correlation matrix to avoid over-smoothing on the features. We also explored several alternative label graph modules to demonstrate the effectiveness of our A-GCN. Extensive experiments on MS-COCO and Fashion550K show that our A-GCN achieves superior performance to several state-of-the-art methods.

\bibliography{reference}
\bibliographystyle{aaai}
\end{document}